\documentclass{article} % For LaTeX2e
\usepackage{main,times}

\usepackage{amsmath,amsfonts,bm}

\def\eqref#1{equation~\ref{#1}}
\def\1{\bm{1}}

\DeclareMathAlphabet{\mathsfit}{\encodingdefault}{\sfdefault}{m}{sl}
\SetMathAlphabet{\mathsfit}{bold}{\encodingdefault}{\sfdefault}{bx}{n}

\usepackage{hyperref}
\usepackage{url}
\usepackage{graphicx}
\usepackage{booktabs}
\usepackage{multirow}
\usepackage[table]{xcolor}
\usepackage{array}
\usepackage{tabularx}
\usepackage[table]{xcolor}
\usepackage{wrapfig}
\usepackage{xcolor}
\usepackage{array}
\usepackage{colortbl}
\usepackage[most]{tcolorbox}

\definecolor{AgentLoopColor}{HTML}{9A6B00}      % 深金黄色
\definecolor{SkillEvolutionColor}{HTML}{4567A5} % 深蓝色  
\definecolor{ActionToolColor}{HTML}{4F7D35}     % 深绿色

\definecolor{TableHeader}{HTML}{F38181}
\definecolor{TableGroupOne}{HTML}{FCE38A}
\definecolor{TableGroupTwo}{HTML}{EAFFD0}
\definecolor{TableOurs}{HTML}{95E1D3}
\definecolor{RBObsInk}{HTML}{244A68}
\definecolor{RBObsHeader}{HTML}{E8EFF6}
\definecolor{RBObsStripe}{HTML}{F5F8FC}
\definecolor{RBObsRule}{HTML}{A9BACB}
\definecolor{tableheader}{RGB}{232,238,245}
\definecolor{tablerow}{RGB}{247,249,252}
\definecolor{tableline}{RGB}{150,165,180}

\title{RoboBridge: A Self-Evolving Embodied Agent Framework for Sim-to-Real Transfer}

\author{%
\parbox[t]{\dimexpr\textwidth-2\tabcolsep\relax}{%
\centering
\bfseries
Chenxi Li$^{1,2}$ \quad
Zhangrui Zhao$^{1,3}$ \quad
Rui Li$^{1,4}$ \quad
Yuan Gao$^{5}$ \\[0.3em]
Kehui Liu$^{1}$ \quad
Jiarui Li$^{1,6}$ \quad
Dong Wang$^{1}$ \quad
Tong Si$^{5}$ \\[0.3em]
Minting Pan$^{1}$ \quad
Wanli Ouyang$^{1}$ \quad
Dongzhan Zhou$^{1}$\thanks{%
Corresponding author: \texttt{zhoudongzhan@pjlab.org.cn}.%
} \\[0.8em]
\normalfont
$^{1}$\,Shanghai Artificial Intelligence Laboratory \\
$^{2}$\,Zhejiang University \\
$^{3}$\,Beihang University \\
$^{4}$\,Harbin Institute of Technology \\
$^{5}$\,Shenzhen Institutes of Advanced Technology \\
$^{6}$\,Shanghai Jiao Tong University
}%
}

\iclrfinalcopy

\begin{document}

\maketitle

% Remove header text and the horizontal rule
\fancyhead{}
\renewcommand{\headrulewidth}{0pt}

\begin{abstract}
%Sim-to-real adaptation remains a major challenge for embodied agents. End-to-end vision-language-action policies provide strong manipulation capabilities, but adapting them to new tasks and physical environments often requires additional demonstrations, policy optimization, or task-specific engineering. Existing embodied agents can organize behavior through tools and reusable procedures, yet they lack a controlled mechanism for revising transferred skills from target-environment feedback.
%We propose RoboBridge, a framework that treats sim-to-real transfer as the continued adaptation of executable task skills. The agent represents task knowledge as procedures connecting task intent, observations, tool operations, and outcome verification. Interaction feedback is used to generate candidate skill revisions, which are evaluated before being persisted or rejected. A pretrained vision-language-action policy is exposed as a reusable action tool and enhanced with inference-time guidance, enabling fine-grained execution without retraining the underlying policy. Skills evolved in simulators are transferred through aligned task semantics, observations, tools, and verification feedback, and can continue to adapt to real-robot tasks. We evaluate the framework on LIBERO-PRO and corresponding physical tasks, studying both skill evolution and post-transfer adaptation. Our framework provides a route from one-shot policy deployment to continual procedural learning across environments.

A key challenge in bringing embodied intelligence into the real world is transferring capabilities from simulation to reality and enabling agents to continually adapt after deployment. End-to-end vision-language-action policies provide strong manipulation capabilities, but their transfer to physical environments typically relies on calibrating simulated visual and dynamical conditions, collecting additional target-domain demonstrations, and optimizing the policy through further training. Tool-using embodied agents offer flexible task orchestration, yet existing systems primarily emphasize task execution and experience reuse within a given environment, with limited support for transferring procedural knowledge and continuously adapting it across simulation and reality. We propose RoboBridge, a framework that treats sim-to-real transfer as the continued adaptation of executable task skills. The agent represents task knowledge as procedures connecting task intent, observations, tool operations, and outcome verification. Interaction feedback is used to generate candidate skill revisions, which are evaluated before being persisted or rejected. A pretrained vision-language-action policy is exposed as a reusable action tool and enhanced with inference-time guidance, enabling fine-grained execution without retraining the underlying policy. RoboBridge grounds transferable skills in task semantics and interaction interfaces shared across simulation and reality. This representation preserves reusable task structure while allowing environment-dependent operations to be selectively revised through real-world execution feedback. We evaluate the framework on LIBERO-PRO and corresponding physical tasks, studying both skill evolution and post-transfer adaptation. Our framework provides a route from one-shot policy deployment to continual procedural learning across environments.
\end{abstract}

\section{Introduction}

\begin{figure}[t]
    \centering
    \includegraphics[width=1\linewidth]{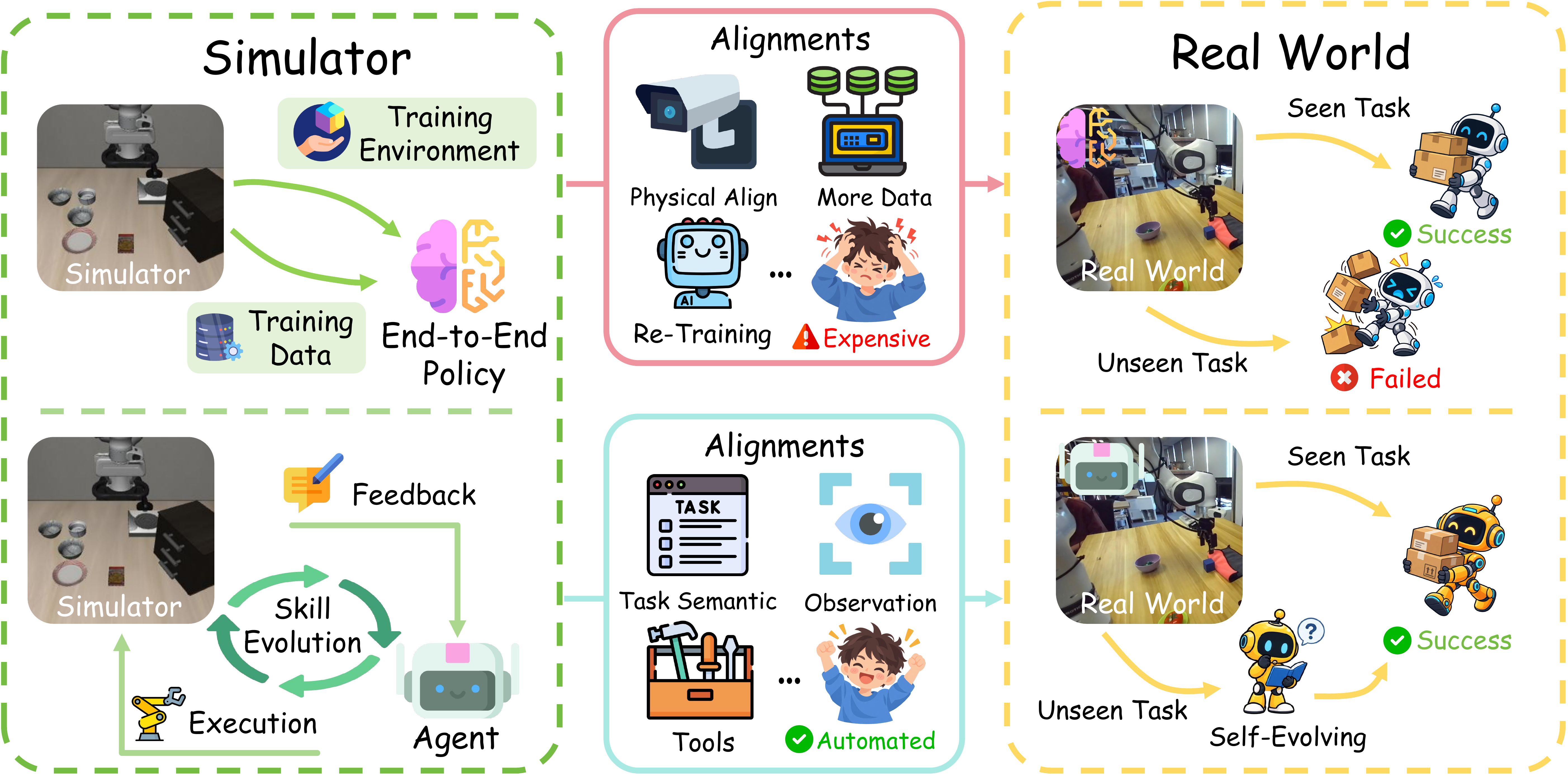}
    \caption{\textbf{Sim-to-real transfer with end-to-end models and RoboBridge.} The policy-centered pathway adapts an end-to-end model for real-world execution through physical alignment, additional target-domain data, and retraining. \textbf{RoboBridge} evolves procedural skills through a feedback-driven execution loop in simulation and transfers them through aligned task semantics and compatible observation and tool interfaces. After deployment, real-world execution feedback drives further skill evolution, refining the transferred procedures for the target environment.}
    \label{fig:teaser}
\end{figure}

Reliable sim-to-real transfer requires robotic systems to preserve task performance across differences in visual observations and physical dynamics \citep{zhao2020sim2real}.
End-to-end vision-language-action models provide fine-grained manipulation capabilities by mapping visual observations and language instructions to action sequences \citep{kim2024openvla,black2024pi_0}. The policy-centered pathway illustrated in Figure~\ref{fig:teaser} includes adapting simulation conditions using real-world experience \citep{chebotar2018closing}. For pretrained VLA models, adaptation to new robotic settings commonly involves target-domain demonstrations and policy fine-tuning \citep{kim2025oft}. In this formulation, task knowledge is largely embedded in model parameters and learned action distributions. This makes it difficult to isolate a reusable task procedure and selectively revise it when execution conditions change, tying procedural adaptation closely to policy optimization.

Tool-using embodied agents offer a way to make task procedures explicit \citep{singh2022progprompt,rana2023sayplan}. An agent harness provides the observation and tool interfaces through which the agent organizes subtasks, invokes operations, and verifies their outcomes \citep{liu2026guava}. Successful procedures can then be retained as skills for subsequent execution \citep{wang2023voyager,lu2026aspire}. Feedback-based memory updates and skill refinement extend this reuse across interactions \citep{shinn2023reflexion,wang2026self}, and recent robotic systems further incorporate repair validation and  validation-gated skill updates \citep{lu2026aspire,ding2026zetta}. Sim-to-real deployment requires these mechanisms to govern which parts of a transferred procedure remain applicable and how the others should be refined using target-environment feedback. Meanwhile, task decomposition and tool selection still require an underlying capability to execute continuous, fine-grained manipulation \citep{yu2023language}. Recent harnesses address this execution requirement by exposing frozen VLA policies as reusable manipulation primitives \citep{zhang2026harness}. This perspective motivates a framework that combines revisable task knowledge with the manipulation capabilities of pretrained policies.

We propose RoboBridge, a self-evolving embodied agent framework for sim-to-real transfer of procedural skills. RoboBridge incorporates a pretrained VLA policy as a reusable action tool within an agent harness. The policy supplies short manipulation segments, while the agent determines how these operations are organized, when they should be invoked, and whether their outcomes satisfy the task. Task knowledge is stored as procedural skills connecting intent, observations, tool operations, and outcome verification. This organization preserves pretrained execution capabilities while exposing task procedures for revision through interaction. As illustrated in Figure~\ref{fig:teaser}, skills evolved in simulation are reused in reality through aligned task semantics and interaction interfaces, with action tools instantiated for each environment. Real-world execution feedback then supports continued skill adaptation. Effective transfer requires both reliable revision of the transferred procedures and accurate execution of their local operations.

 Reliable execution presents a complementary challenge: even a suitable task procedure can fail when a frozen policy produces actions that do not meet the current subtask requirements \citep{liu2026vls}. Prior work has explored adapting pretrained policy behavior during inference through external guidance or verification \citep{wang2024steering,wu2025foresight,liu2026vls}. RoboBridge therefore augments the policy tool with inference-time guidance. The agent determines whether guidance is needed and configures a reward specification from the subtask objective, execution constraints, and observed feedback. This reward steers action generation toward the intended operation while keeping the policy parameters fixed. Skill evolution thus adjusts the procedure across episodes, while inference-time guidance adapts action generation within an invocation.

Our contributions are summarized as follows:
\begin{enumerate}
    \item \textbf{A self-evolving framework for procedural sim-to-real transfer.}
    We integrate pretrained action capabilities with explicit task skills, enabling procedures evolved in simulation to be reused through aligned interaction interfaces and further adapted from real-world feedback.

    \item \textbf{Controlled skill evolution.}
    We develop a feedback-driven process for adding, revising, and deleting procedural knowledge, with validation governing which updates are persisted and which are rejected or rolled back.

    \item \textbf{Adaptive policy-tool orchestration.}
    We enable the agent to determine when to invoke the pretrained policy, when to activate inference-time guidance, and how to configure subtask rewards, supporting context-dependent execution with frozen policy parameters.
\end{enumerate}
\section{Related Work}

\subsection{End-to-End Embodied Models}

End-to-end embodied models learn to map visual observations and language instructions directly to robot actions. Early systems such as RT-1 \citep{brohan2022rt} and RT-2 \citep{brohan2023rt} established this paradigm, while later models expanded its scope through embodied multimodal reasoning, multi-robot datasets, and generalist policies, including PaLM-E \citep{driess2023palm}, Open X-Embodiment \citep{o2024open}, Octo \citep{team2024octo}, OpenVLA \citep{kim2024openvla}, and $\pi_{0}$ \citep{black2024pi_0}. These models improve generalization by increasing the diversity of training tasks, observations, robot embodiments, and demonstrations, but their behavior remains primarily encoded in a fixed policy. To address the gap between simulation and physical deployment, existing approaches increase the diversity of simulated observations \citep{tobin2017domain}, perturb simulated dynamics \citep{peng2018sim}, or adapt the simulator distribution using informative variations and posterior estimation \citep{mehta2020active, muratore2022neural}. Recent data-centric methods further scale simulated experiences and adaptation data to improve transfer across tasks and robot embodiments~\citep{liu2025dexscale}, while inference-time steering adjusts a frozen policy during execution~\citep{liu2026vls}. These approaches use a single end-to-end policy to map observations and instructions to actions, leaving adaptation coupled to the policy itself. Our framework instead uses an agent to organize reusable skills and invoke action tools, so the transferred procedure can be revised in the target environment without retraining the entire policy.

\subsection{Embodied Agents}
Embodied-agent research introduces explicit reasoning, tool use, and feedback into robotic execution. Early systems grounded language in executable affordances and used environment feedback for closed-loop planning \citep{ahn2022can, huang2022inner}, while program-based approaches synthesized control logic or composed spatial value functions for manipulation \citep{liang2023code, huang2023voxposer}. Subsequent work explored skill bootstrapping, lifelong skill libraries, and open-vocabulary adaptation through interaction and memory \citep{zhang2023bootstrap, tziafas2024lifelong, zhao2025vla}. Recent systems have further studied workflow reasoning, self-directed skill acquisition, feedback-based repair, and continual adaptation in simulated or physical environments \citep{lu2026aspire, ding2026zetta, wang2026self, nie2026teach, liu2026guava, lei2026ace, zhang2026playful}. These studies show that explicit procedures, persistent memory, and execution feedback can extend embodied reasoning beyond direct policy execution. Their primary focus, however, varies between workflow planning, skill discovery, memory expansion, and runtime self-improvement. Our work focuses on the migration process itself: a skill is treated as a revisable task hypothesis, and feedback from the target environment determines which parts of the transferred procedure should be retained, modified, or removed.
\section{Method}

\begin{figure}[t]
    \centering
    \includegraphics[width=1\linewidth]{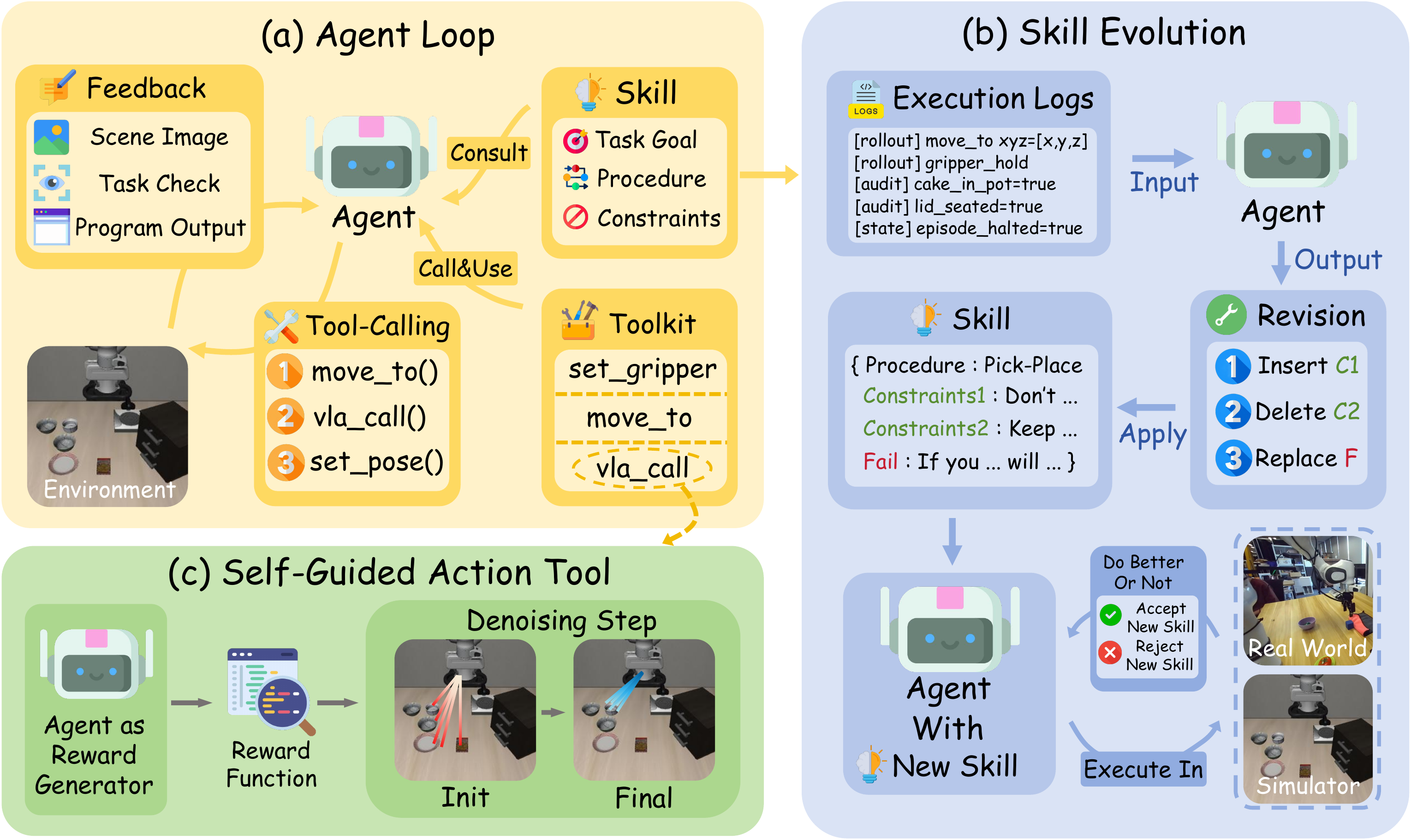}
    \caption{\textbf{RoboBridge framework.}
    \textcolor{AgentLoopColor}{\textbf{(a) Agent loop:}}
    The agent consults the active skill, invokes manipulation tools, and uses observations and execution feedback to guide subsequent actions. \textcolor{SkillEvolutionColor}{\textbf{(b) Skill evolution:}} Execution logs inform candidate insertions, deletions, and replacements in the skill. Candidates are evaluated through task execution in the current environment, with validated improvements retained and unsuccessful updates rejected.
    \textcolor{ActionToolColor}{\textbf{(c) Self-guided action tool:}} When needed, the agent configures a reward function to guide action refinement during denoising, adapting execution to the current subtask while keeping the pretrained model parameters fixed.}
    \label{fig:method_main}
\end{figure}

 RoboBridge integrates skill-guided execution, inference-time action guidance, and feedback-driven skill evolution into the closed loop illustrated in Figure~\ref{fig:method_main}. During task execution, the agent consults the active skill and current observations to select tool operations, then uses task checks and program outputs to guide subsequent decisions. When invoking the pretrained end-to-end action model, the agent can construct a reward function to steer action refinement during denoising while keeping model parameters fixed. Execution logs from the rollout provide evidence for adding, deleting, or replacing skill content. Candidate skills are evaluated through task execution, and validated improvements are retained to guide subsequent rollouts while unsuccessful revisions are rejected. This loop allows execution experience to improve reusable procedures and updated procedures to guide future actions. With aligned task semantics and compatible observation and tool interfaces, skills evolved in the simulator can initialize real-world execution and continue to adapt through physical feedback.

\subsection{Problem Formulation and Agentic Execution Loop}

  \textbf{Task Setup and Skill Evolution.}
  We consider language-conditioned manipulation in a source simulator, LIBERO-PRO \citep{zhou2510libero}, and corresponding real-robot tasks in a target environment. Each task is specified by an instruction $\ell$ and a completion condition $G$. The agent maintains an
  editable skill $S$ that connects observations to tool operations and verification conditions, providing procedural knowledge for task execution.

  Skill evolution uses interaction feedback to revise this knowledge. In either environment, a task rollout $\tau_k$ informs the validated update process $\mathcal{U}$. Once evolution in simulation is complete, the resulting skill initializes
  adaptation in the physical environment through an interface mapping $\Phi$:
  \begin{equation}
  S_{k+1} = \mathcal{U}(S_k, \tau_k),
  \qquad
  S^{r}_{0} = \Phi(S^{s}_{K}),
  \end{equation}
  where $k$ indexes evolution steps, $s$ and $r$ denote simulation and reality, and $K$ is the number of source-side evolution steps. The update process retains the current skill when a candidate fails validation. The mapping $\Phi$ captures the
  alignment of task semantics, observation formats, and tool interfaces across environments. The transferred skill continues to evolve through $\mathcal{U}$ using real-world execution feedback. Our objective is to improve task success with limited
  target-environment interaction while preserving valid procedures and revising those affected by environmental differences.

  \textbf{Agentic Execution Loop.}
  Each rollout consists of successive interactions between a high-level planner $\Pi$ and a unified tool interface $\mathcal{P}$. At turn $t$, the planner receives an observation $o_t$ containing the structured scene description, camera observations,
  robot state, and observation freshness and uncertainty. Together with instruction $\ell$, previous execution feedback $f_{t-1}$, and the active skill $S_k$, this information determines the next tool invocation:
  \begin{equation}
  c_t = \Pi(\ell, o_t, f_{t-1}, S_k),
  \qquad
  (o_{t+1}, f_t) = \mathcal{E}(o_t, c_t),
  \end{equation}
  where $c_t$ is a structured call selected through $\mathcal{P}$, and $\mathcal{E}$ denotes tool execution and observation refresh. The interface provides deterministic manipulation primitives, including gripper and pose controls, policy-based action
  tools, and task termination.

  Each turn contains at most one physical operation. Once the operation completes or reports a failure, the updated observation and feedback return to the planner, allowing subsequent decisions to reflect the actual execution outcome. The loop ends
  when $G$ is satisfied, the agent explicitly terminates, or the episode budget is exhausted. The resulting trajectory records observations, tool calls, and feedback as evidence for subsequent skill updates, linking feedback-guided execution to skill
  evolution across rollouts.

\subsection{Embodied Feedback-Driven Skill Evolution}
\label{sec:skill-evolution}

As the experience-bearing unit of embodied-agent self-evolution, each skill is organized as an executable hypothesis about how a task should be performed. It captures the relation between changes in the observed scene and the agent's subsequent behavior: given the current observation and feedback from the previous operation, it specifies how the next operation should be selected and executed. At the same time, a skill is not tied to a particular object identity. It is organized around task-relevant states and manipulation affordances. For example, a grasping skill is not intended for a single object such as a mug, but for different targets that share similar manipulation requirements. This organization separates task intent from its physical realization, allowing the skill to preserve reusable behavioral structure as the environment changes.

Effective skill evolution requires more than adding new experience; it requires continual revision of existing knowledge. To prevent erroneous experience from permanently contaminating the skill memory and to avoid uncontrolled append-only accumulation, a skill must be tested through embodied feedback during task execution. The system collects scene observations, tool outcomes, intermediate verification signals, and the final task result throughout each rollout. Together, these signals provide multiple forms of evidence about the current skill. Successful trajectories can indicate that an operation and its surrounding conditions are reusable, whereas failed trajectories may reveal an unsuitable operation, an incorrect execution order, a missing transition, or a verification condition that no longer holds in the target environment. After a rollout, the agent integrates this evidence to propose a local revision of the current skill. The revision may add a missing operation, replace an unreliable transition, or remove behavior that is no longer supported, while preserving unaffected parts of the procedure.

To distinguish evidence-supported improvements from unverified behavioral hypotheses, each candidate skill is evaluated against the current skill under comparable task conditions. A candidate is consolidated as a new skill version only when it remains valid and improves task execution. If it fails to improve performance or introduces invalid behavior, the current skill remains unchanged and the candidate is excluded from persistent knowledge. Skill evolution is therefore not a simple accumulation of experience, but a feedback-driven process in which behavioral evidence is tested, selectively incorporated, and gradually consolidated into reusable task knowledge.

\subsection{From a Frozen Policy to a Self-Guided Action Tool}

RoboBridge integrates the fine-grained manipulation capabilities of a pretrained end-to-end action model with the agent's explicit task procedures. The agent uses the active skill and multimodal context to specify a local operation, which the model translates into a short action segment. Operation selection and outcome verification remain explicit in the agent's execution loop. This organization allows procedural knowledge to be revised independently of the model parameters while retaining pretrained manipulation capabilities.

The central execution challenge is to align the model's generated actions with the requirements of the selected operation. A valid task procedure specifies the intended outcome, but a frozen model can still produce unsuitable behavior under different observations and physical conditions. We address this discrepancy through inference-time guidance configured by the agent. When guidance is needed, the agent uses task context and execution feedback to construct a reward that expresses subtask objectives and execution constraints. Let $\mathbf{A}_t^{(j)}$ denote the candidate action sequence at denoising step $j$ during the tool invocation at turn $t$. The reward-guided correction is expressed as
\begin{equation}
    \mathbf{A}_{t,\mathrm{guided}}^{(j)}
    =
    \mathbf{A}_t^{(j)}
    +
    \eta_{t,j}
    \nabla_{\mathbf{A}_t^{(j)}}
    r_t\bigl(\mathbf{A}_t^{(j)};o_t,u_t\bigr),
    \label{eq:reward-guided-action}
\end{equation}
where $o_t$ is the current observation, $u_t$ specifies the local subtask, and $r_t$ is the agent-configured reward function. The coefficient $\eta_{t,j}\geq 0$ controls the guidance strength, with $\eta_{t,j}=0$ corresponding to no reward-guided correction. The gradient is taken with respect to the candidate actions, steering their refinement toward higher reward during denoising while keeping the pretrained model parameters fixed. This mechanism connects the agent's task reasoning to action generation, enabling behavioral adaptation within an individual tool invocation.

This mechanism complements skill evolution at the level of individual operations. Skill evolution revises the procedural knowledge governing operation selection, ordering, and verification across episodes, while inference-time guidance adjusts the execution of the selected operation under the current conditions. Execution feedback informs both subsequent planning and further skill revision, linking immediate action adjustment to persistent changes in task knowledge.

\section{Experiments}
\label{sec:experiment}

We organize our experiments around two deployment stages. In the simulation stage, the agent repeatedly interacts with LIBERO-PRO to evolve task skills and evaluates the resulting procedures. In the transfer stage, the skills obtained in simulation are initialized on corresponding real-robot tasks and further adapted from feedback collected in the target environment.

For each task in both the simulated and real-world settings, we follow the Embodied Feedback-Driven Skill Evolution process described in Section \ref{sec:skill-evolution} to obtain the skill used for subsequent evaluation. The agent gathers evidence from multiple training episodes and uses it to generate candidate skills. After evaluation, an accepted candidate is persisted as a new skill version for subsequent deployment.

Section \ref{sec:experimental-setup} describes the common experimental configuration and skill evolution protocol. Section \ref{sec:libero-pro-experiments} reports the results on the LIBERO-PRO benchmark, and Section \ref{sec:real-robot-experiments} evaluates skill transfer and continued adaptation in the real-robot environment.

\subsection{Experimental Setup}
\label{sec:experimental-setup}
 \begin{table}[t]
      \centering
      \begingroup

      \hypersetup{hidelinks}

      \caption{Success rates (\%) on
      LIBERO-PRO~\citep{zhou2510libero} across SPATIAL,
      OBJECT, GOAL, and LIBERO-10 under instruction-redirection (T)
      and position-swap (S) perturbations.
      Our evaluation uses 100 trials per suite--perturbation pair
      (10 tasks $\times$ 10 seeds).
      Spat and Long denote SPATIAL and LIBERO-10, respectively.
      Dashes indicate unreported results.
      Overall scores for prior methods are reproduced as reported,
      with differing suite coverage where results are unavailable.
      Bold indicates the highest reported score in each column.}
      \label{tab:libero-main}

      % Space between the caption and the table.
      \vspace{6pt}

      \small
      \setlength{\tabcolsep}{1.25pt}
      \renewcommand{\arraystretch}{1.15}
      \setlength{\extrarowheight}{1pt}

      % Keep row backgrounds continuous around horizontal rules.
      \setlength{\aboverulesep}{0pt}
      \setlength{\belowrulesep}{0pt}

      \newcommand{\tblcite}[1]{{\footnotesize\citep{#1}}}

      % X columns absorb the extra width within colored cells.
      % The final column uses its natural width for "Overall".
      \begin{tabularx}{\linewidth}{
          l
          *{8}{>{\centering\arraybackslash}X}
          c
      }
          \toprule

          \rowcolor{TableHeader!20}
          & \multicolumn{4}{c}{Instruction redirection (T)}
          & \multicolumn{4}{c}{Position swap (S)}
          & \\

          \rowcolor{TableHeader!20}
          \multirow{-2}{*}{Method}
          & Spat & Object & Goal & Long
          & Spat & Object & Goal & Long
          & \multirow{-2}{*}{Overall} \\

          \midrule

          \rowcolor{TableGroupOne!20}
          OpenVLA~\tblcite{kim2024openvla}
          & 0.0 & 0.0 & 0.0 & 0.0
          & 0.0 & 0.0 & 0.0 & 0.0
          & 0.0 \\

          \rowcolor{TableGroupOne!20}
          $\pi_{0}$~\tblcite{black2024pi_0}
          & 0.0 & 0.0 & 0.0 & 0.0
          & 0.0 & 2.0 & 0.0 & 0.0
          & 0.3 \\

          \rowcolor{TableGroupOne!20}
           $\pi_{0.5}$~\tblcite{intelligence2025pi_}
          & 1.0 & 1.0 & 2.0 & 1.0
          & 20.0 & 17.0 & 38.0 & 8.0
          & 11.0 \\

          \rowcolor{TableGroupOne!20}
          MolmoAct~\tblcite{lee2025molmoact}
          & 0.0 & 0.0 & 0.0 & 6.0
          & 0.0 & 6.0 & 0.0 & 0.0
          & 1.5 \\

          \rowcolor{TableGroupOne!20}
          NORA~\tblcite{hung2025nora}
          & 0.0 & 0.0 & 0.0 & 0.0
          & 0.0 & 0.0 & 0.0 & 0.0
          & 0.0 \\

          \rowcolor{TableGroupOne!20}
          X-VLA~\tblcite{zheng2025xvla}
          & 0.0 & 8.0 & 9.0 & 10.0
          & 0.0 & 2.0 & 1.0 & 0.0
          & 3.8 \\

          \rowcolor{TableGroupOne!20}
          AtomVLA~\tblcite{sun2026atomvla}
          & 1.0 & 0.0 & 11.0 & 9.0
          & 16.0 & 10.0 & 2.0 & 1.0
          & 6.3 \\

          \rowcolor{TableGroupOne!20}
          $\pi_{\mathrm{RLinf}}$~\tblcite{yu2025rlinf}
          & 42.0 & 71.0 & 45.0 & 49.0
          & 59.0 & 78.0 & 42.0 & 14.0
          & 50.0 \\

          \midrule

          \rowcolor{TableGroupTwo!20}
          CaP-X~\tblcite{fu2026capx}
          & 14.0 & 18.0 & 17.0 & ---
          & 12.0 & 22.0 & 26.0 & ---
          & 18.2 \\

          \rowcolor{TableGroupTwo!20}
          RATs~\tblcite{zhang2026playful}
          & 31.0 & 63.0 & 36.0 & ---
          & 29.0 & 61.0 & 43.0 & ---
          & 43.8 \\

          \rowcolor{TableGroupTwo!20}
          ASPIRE~\tblcite{lu2026aspire}
          & 60.0 & \textbf{95.0} & 45.0 & ---
          & 51.0 & \textbf{98.0} & \textbf{81.0} & ---
          & 71.6 \\

          \rowcolor{TableGroupTwo!20}
          Harness VLA (Codex)~\tblcite{zhang2026harness}
          & 81.0 & 94.0 & 75.0 & 52.0
          & 69.0 & 91.0 & 66.0 & 49.0
          & 72.1 \\

          \midrule

          \rowcolor{TableOurs!20}
          \textbf{Ours (Codex)}
          & \textbf{83.0} & 94.0 & \textbf{80.0} & \textbf{55.0}
          & \textbf{73.0} & 95.0 & 65.0 & \textbf{50.0}
          & \textbf{74.4} \\

          \bottomrule
      \end{tabularx}

      \endgroup
  \end{table}

We evolve skills and initially evaluate the framework in LIBERO-PRO, then transfer the resulting skills to a real-robot environment for further evolution and evaluation. We use Codex with GPT-5.5 as the coding agent. Across all experiments, the agent operates through a unified observation and tool interface. The planner can combine deterministic operation tools with policy-based action tools and receives an updated observation and execution feedback after each operation.

The end-to-end action primitive is instantiated with different checkpoints for simulation and real-robot execution while preserving the same action interface. Specifically, we use the RLinf-released \texttt{pi05\_libero130\_fullshot} $\pi_{0.5}$-SFT checkpoint for LIBERO-PRO and the OpenPI-released $\pi_{0.5}$-DROID checkpoint for the real-robot tasks.

For each task unit in LIBERO-PRO, we perform two rounds of skill evolution, with five training seeds used in each round, to obtain the skill version used for evaluation. The detailed allocation of training, validation, and test seeds is provided in Appendix~\ref{app:liberopro-evaluation}. In the real-robot environment, we perform one round of skill evolution at the end of each episode.

\subsection{Simulation Evaluation}
\label{sec:libero-pro-experiments}
We deploy the skills produced by skill evolution and evaluate our framework on previously unseen seeds. This evaluation tests whether the framework can reuse strategies for organizing actions learned on the training seeds under new task conditions. Specifically, we examine whether the evolved skills help the agent maintain robustness to instruction-redirection (T) and position-swap (S) perturbations in LIBERO-PRO.

Table \ref{tab:libero-main} summarizes the results on LIBERO-PRO across four task suites, SPATIAL, OBJECT, GOAL, and LIBERO-10, under both T and S perturbations. Existing end-to-end VLA models exhibit marked performance degradation under these distribution shifts. Among prior methods evaluated under identical or similar experimental settings, Harness VLA is the strongest reported baseline, with an overall success rate of $72.1$\%. Our framework achieves an overall success rate of $74.4$\% when using Codex, exceeding Harness VLA by $2.3$ percentage points in the main comparison and by $5$ percentage points on GOAL-T. Under our experimental setup, directly using $\pi_{\mathrm{RLinf}}$ yields an overall success rate of $50.0$\%, indicating that the improvement does not arise solely from the underlying end-to-end policy. The gains across both generalization settings suggest that our framework evolves reusable experience through interaction feedback and a fixed library of action primitives, supporting rapid transfer to similar scenes or tasks with the same semantics.
\begin{figure}[t]
    \centering
    \includegraphics[width=0.85\linewidth]{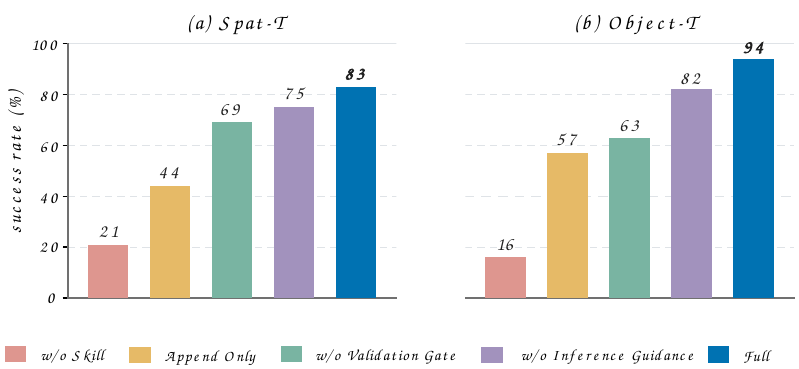}
    \caption{Ablation study on SPATIAL-T and OBJECT-T in LIBERO-PRO.
  We compare the full framework with four variants:
  without skill guidance, with append-only updates that
  exclude revision and deletion, without validation before
  skill persistence, and without inference-time policy guidance.
  Bars report success rates (\%), with the full framework
  achieving 83\% and 94\% on the two suites, respectively.}
    \label{fig:ablation}
\end{figure}
\vspace{-5pt}

%We further investigate the contributions of four key components: the full evolution of embodied Skills, deletion and revision during Skill updates, validation gating before Skill persistence, and inference-time guidance for the end-to-end model tool. Figure \ref{fig:ablation} summarizes the ablation results on SPATIAL-T and OBJECT-T. Although the framework retains some baseline capability without effective Skill guidance, incorporating the fully evolved Skills improves success rates by $62$\% and $78$\% on the two suites, respectively. Removing deletion and revision and allowing only append operations reduces success rates by $39$\% and $36$\%, respectively. With append-only updates, immature experience accumulated during early evolution can continue to misguide the agent, while increasingly verbose Skill content reduces the precision and effectiveness of guidance. The validation gate evaluates textual Skill updates through their effects on task performance before allowing them to persist, helping prevent negative transfer and improving success rates by $14$\% and $21$\% on the two suites, respectively. With the full Skill Evolution process in place, inference-time guidance for the end-to-end model tool further reduces errors in fine-grained manipulation, yielding additional improvements of $8$\% and $12$\%, respectively.

We further investigate the contributions of four components: procedural skill guidance, deletion and revision during skill updates, validation before skill persistence, and inference-time guidance for the pretrained end-to-end action model. Figure~\ref{fig:ablation} summarizes the results on SPATIAL-T and OBJECT-T, where the full framework achieves success rates of 83\% and 94\%, respectively. Without skill guidance, success rates drop to 21\% and 16\%, corresponding to decreases of 62 and 78 percentage points. Restricting skill evolution to append-only updates reduces success rates to 44\% and 57\%, which are 39 and 37 percentage points below the full framework. These results highlight the importance of revising and removing existing procedural knowledge during skill evolution.
Removing the validation gate lowers success rates to 69\% and 63\%, corresponding to decreases of 14 and 31 percentage points. This result supports evaluating candidate skill updates through task execution before incorporating them into persistent skill memory. With the full skill evolution process retained, disabling inference-time guidance reduces success rates to 75\% and 82\%. Activating guidance therefore provides additional gains of 8 and 12 percentage points, respectively. Together, these ablations show that validated skill evolution and inference-time action guidance make complementary contributions to task performance.

\subsection{Real-Robot Evaluation}
\label{sec:real-robot-experiments}

We evaluate whether task skills evolved in simulation can be transferred directly to real-robot settings and further improved through feedback from physical execution. This evaluation examines whether simulation experience can facilitate adaptation to real-robot tasks and reduce the effort required for debugging. To support reuse across environments, we encapsulate the real robot's perception, calibration, and control pipelines behind a unified interface, exposing observation specifications and compatible tool interfaces aligned with those used in simulation. We investigate three questions: (1) whether skills evolved in simulation can be applied directly to real-robot tasks with the same or similar task semantics; (2) whether transferred skills can continue to evolve through real-world interaction and improve task performance; and (3) whether simulation skills reduce the exploration and debugging effort required for learning and evolution in the physical environment.

\begin{figure}[t]
    \centering
    \includegraphics[width=0.9\linewidth]{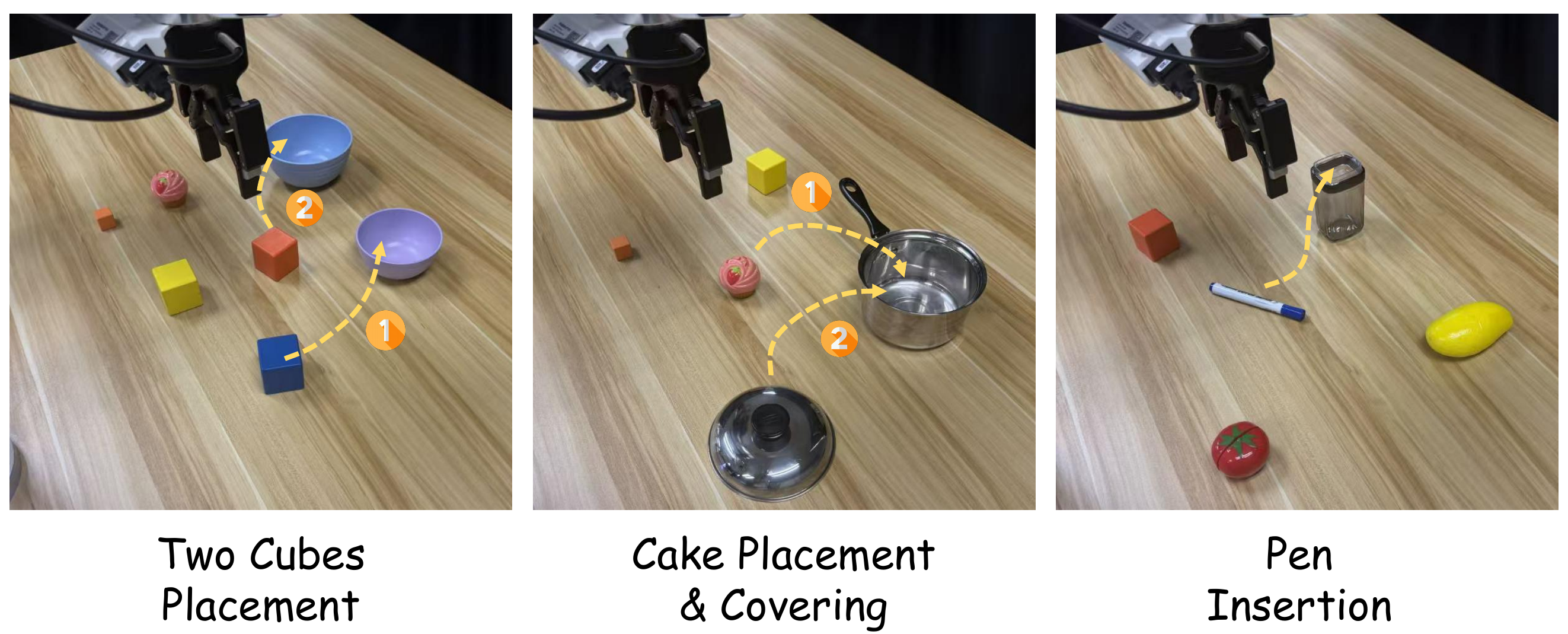}
    \caption{\textbf{Real-robot evaluation tasks.}
  Left: Placing two cubes into their respective target bowls.
  Middle: Placing a cake into a pot and covering it with the lid.
  Right: Inserting a pen into a holder.}
    \label{fig:real_robot}
\end{figure}

\begin{figure}[t]
    \centering
    \includegraphics[width=1\linewidth]{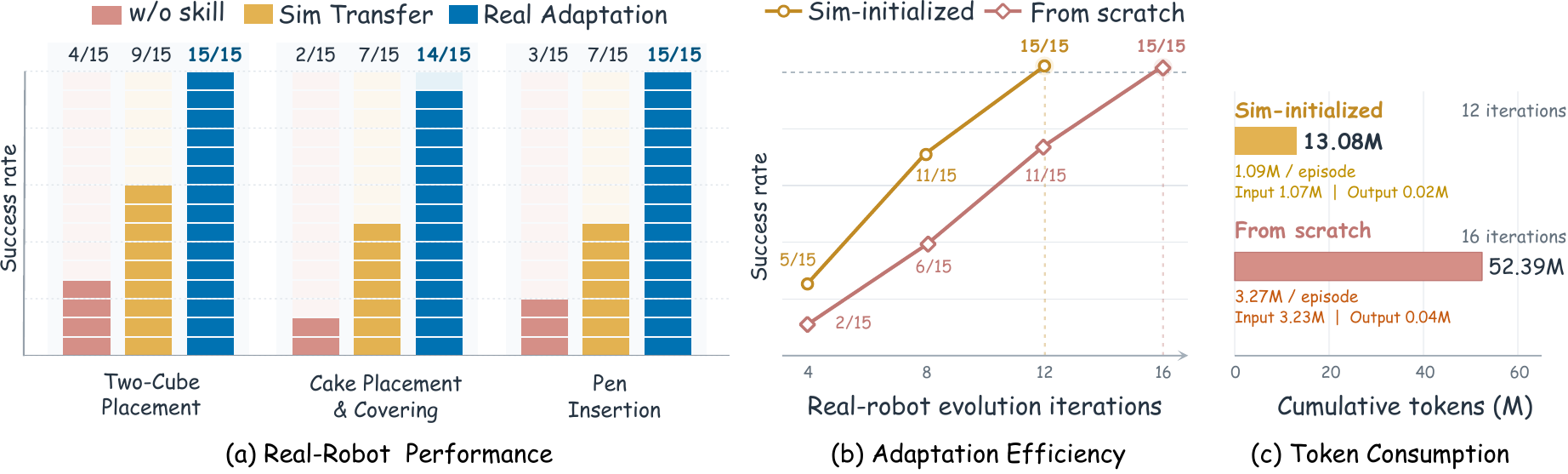}
    \caption{\textbf{Real-robot performance and adaptation efficiency.}
  \textbf{(a) Real-robot performance:} Success counts out of $15$ trials on three tasks without skill guidance, with direct transfer of simulation-evolved skills, and after $10$ episodes of real-robot skill evolution.
  \textbf{(b) Adaptation efficiency:} Under a separate experimental setting on the cake placement and covering task, simulation-initialized skill evolution reaches $15/15$ after $12$ iterations, compared with $16$ iterations from an empty skill state. Each iteration consists of one physical execution episode followed by one skill update.
  \textbf{(c) Token consumption:}
  Cumulative token usage for the same task and setting as (b), measured at the respective $12$- and $16$-iteration endpoints. Annotations report mean total, input, and output tokens per evolution episode. M denotes million tokens.}
    \label{fig:real_experiment}
\end{figure}

As shown in Figure \ref{fig:real_robot}, we design three real-robot tasks whose task semantics and high-level operation procedures correspond to existing simulation tasks, while their specific objects and manipulation modes differ. We first apply the skills evolved in simulation directly to the physical tasks, before performing any real-robot evolution, to measure their transfer value. Figure \ref{fig:real_experiment} (a) shows that, compared with execution without skill guidance, the transferred skills improve the success rates of the three tasks from $4/15$, $2/15$ and $3/15$ to $9/15$, $7/15$ and $7/15$. These results indicate that simulation experience provides effective initial behavioral guidance for real-robot execution.

%We then initialize real-robot adaptation with the transferred skills. For each task, we run $10$ real-robot episodes and perform one round of Skill Evolution after every episode. The resulting execution feedback is used to revise the skills and adapt them to the target environment and its manipulation conditions. Compared with the initial transferred skills before real-robot evolution, the adapted skills further improve task performance, achieving $15/15$, $14/15$, and $15/15$ successful trials on the three tasks, respectively. This result shows that transferred skills can continue to improve through interaction in the physical environment, supporting adaptation after deployment.

We then initialize real-robot adaptation with the transferred skills. For the performance evaluation in
Figure~\ref{fig:real_experiment} (a), we run $10$ real-robot episodes per task and perform one round of skill evolution after every episode. Execution feedback is used to revise the skills for the target environment and its manipulation conditions. The adapted skills achieve $15/15$, $14/15$, and $15/15$ successful trials on the three tasks, respectively. These improvements over direct transfer show that physical interaction can further refine the transferred skills after deployment.
  
%To further quantify the effect of simulation experience on adaptation efficiency, we compare real-robot evolution initialized from simulation skills with real-robot evolution that starts from an empty skill state. As shown in Figure \ref{fig:real_experiment} (b) and (c), reaching the same performance level from scratch requires more than twice as many real-robot iterations as evolution initialized from simulation skills. During the early and middle stages of evolution, the absence of prior experience causes the agent to spend more computation on reasoning, exploration, and repeated trial-and-error. The from-scratch process also frequently reaches the predefined limit on the number of attempts, resulting in a cumulative token consumption nearly four times that of the simulation-initialized process. These observations show that simulation skills provide more than an initial execution prior: they reduce the need to rediscover task procedures during real-robot evolution.

To further quantify the effect of simulation experience on adaptation efficiency, we compare skill  evolution initialized from simulation skills with evolution from an empty skill state. Figure~\ref{fig:real_experiment}(b) shows that simulation-initialized evolution reaches $15/15$ after $12$ iterations, compared with $16$ iterations from scratch, corresponding to a $25\%$ reduction in iterations. At these respective endpoints, cumulative token consumption is $13.08$M and $52.39$M, respectively, as shown in Figure~\ref{fig:real_experiment}(c). Simulation initialization thus reduces cumulative token consumption by approximately $75\%$. Together, these results show that transferred skills reduce both the interaction budget and token consumption needed to reach the same performance level on this task.

Together, these results show that, when task semantics and interaction interfaces are aligned, procedural skills formed in simulation provide reusable initial knowledge for real-robot tasks and can continue to adapt from physical execution feedback. The gains from direct transfer demonstrate the reuse value of simulation experience, while the additional improvements after real-robot evolution show that target-environment feedback can refine the transferred procedures. Compared with evolution from scratch, reusing simulation skills also reduces the number of iterations and the token budget required to reach the same performance level, indicating benefits in both task effectiveness and adaptation efficiency. Combining simulation-time experience accumulation with post-deployment skill evolution therefore provides a more efficient starting point for real-robot adaptation by reducing the interaction and reasoning overhead of physical exploration.

\vspace{-5pt}
\section{Conclusion And Limitations}
\vspace{-5pt}
We presented RoboBridge, a self-evolving embodied agent framework for transferring procedural skills from simulation to reality and adapting them through execution feedback. Reusable, revisable skills guide tool selection and outcome verification, while inference-time guidance adapts action generation without updating the underlying policy. Experiments on LIBERO-PRO and three real-robot tasks show that simulation-evolved skills can provide useful initialization for physical execution and continue to improve through real-world feedback, reducing adaptation effort compared with evolution from scratch. Our current evaluation adopts Codex with GPT-5.5 as the coding-agent configuration. A direction for future work is to explore RoboBridge with a broader range of foundation models and agent harnesses, further examining its applicability across different model capabilities and agent execution frameworks.

\subsection*{AI use statement}

We used OpenAI Codex to assist with translation, manuscript drafting and language editing, and the writing, debugging, modification, and execution of research code. Codex also assisted in organizing experimental results and qualitatively analyzing multimodal execution records to support the interpretation of experimental processes and observed phenomena. Additional uses included literature searches, citation and bibliography preparation, and the creation and editing of scientific figures and plotting code. Generative AI was not used to formulate the core research framework or hypotheses, design the experiments, generate synthetic datasets, clean research datasets, or formulate mathematical claims or proofs.

The authors reviewed the AI-assisted text, translations, and interpretations against the implemented method and experimental evidence. We verified references against the original publications, checked numerical results, token statistics, and figures against the underlying experimental records and logs, and inspected and tested the AI-assisted code. The authors take responsibility for the final content of this work, including its text, claims, references, experimental reporting, and AI-assisted artifacts.

\bibliography{main}
\bibliographystyle{main}

\appendix
\newpage
\section{Planner Design with Decoupled Observation and Execution}
  \label{app:planner-design}

RoboBridge separates observation construction from action selection through an environment adapter. In centralized-perception mode, the adapter supplies an initial observation after reset and refreshes it after each action-tool invocation. The planner receives the task instruction, procedural skills, current observation, and execution feedback, then selects the next tool and its arguments. Segmentation, geometric reconstruction, and observation packaging are handled by the adapter. The planner is responsible for task decomposition, object selection, motion offsets, and outcome verification. This division allows sensing and control implementations to vary across environments while preserving the observation--action--verification loop used by the skills.

In simulation, observation construction begins with a scene inventory initialized from the camera images and task instruction. The inventory provides entity identities and descriptions for subsequent segmentation. After each action, the adapter associates the resulting masks with current metric geometry and extracts measured reference points. Each point retains its entity identifier, source camera, pixel location, and capture step. Missing or unreliable measurements are represented through availability and uncertainty fields. The resulting snapshot describes the scene and robot state; target--destination assignments, grasp offsets, and end-effector goals are determined by the planner. For example, a measured point on an object's upper surface supplies a geometric reference, while the planner determines the approach and grasp positions using the task, skill, and current observations.

Table~\ref{tab:planner-observation} summarizes the simulation observation schema. It specifies the coordinate frame and units alongside entity geometry and robot proprioception. Four fixed image slots contain external and wrist RGB views and their annotated counterparts. The serialized snapshot stores image metadata, while the runtime attaches image content to the planner request. Task language and execution feedback are supplied alongside the snapshot. This organization keeps observed scene properties distinct from the task-dependent decisions made from them.

 \begin{table}[h]
      \centering
      \caption{Structured observation provided to the planner in simulation.
      The snapshot combines observation identity, coordinate conventions,
      external and wrist image metadata, scene entities and their measured
      geometry, robot proprioception, and validity information. Positions
      are expressed in the simulation world frame in metres. Optional
      estimates are included when available. Task instructions and
      execution feedback accompany the snapshot as separate inputs.}
      \label{tab:planner-observation}

      \vspace{8pt}

      \begingroup
      \setlength{\tabcolsep}{8pt}
      \renewcommand{\arraystretch}{1.15}
      \setlength{\extrarowheight}{3pt}
      \setlength{\aboverulesep}{0pt}
      \setlength{\belowrulesep}{0pt}

      % Both columns use vertically centered paragraph cells.
      \renewcommand{\tabularxcolumn}[1]{m{#1}}
      \arrayrulecolor{RBObsRule}

      \begin{tabularx}{\linewidth}{
          >{\raggedright\arraybackslash}m{0.30\linewidth}
          >{\raggedright\arraybackslash}X
      }
          \toprule[0.8pt]

          \rowcolor{RBObsHeader}
          \textcolor{RBObsInk}{\textbf{Field}}
          & \textcolor{RBObsInk}{\textbf{Contents}} \\

          \midrule[0.4pt]

          \textcolor{RBObsInk}{\texttt{observation\_id}}\newline
          \textcolor{RBObsInk}{\texttt{episode\_step}}
          & Unique observation identifier and the corresponding
          action-tool step. \\

          \rowcolor{RBObsStripe}
          \textcolor{RBObsInk}{\texttt{coordinate\_frame}}\newline
          \textcolor{RBObsInk}{\texttt{units}}
          & Reference frame and measurement units:
          simulation world frame and metres. \\

          \textcolor{RBObsInk}{\texttt{images}}
          & External and wrist RGB views and their annotated counterparts,
          with availability, camera, capture step, resolution,
          and content references. \\

          \rowcolor{RBObsStripe}
          \textcolor{RBObsInk}{\texttt{entities}}
          & Entity identifiers, visual labels, visibility, available
          geometric estimates, and measured reference points
          with provenance. \\

          \textcolor{RBObsInk}{\texttt{relations}}
          & Supported relations between entity identifiers. \\

          \rowcolor{RBObsStripe}
          \textcolor{RBObsInk}{\texttt{robot}}
          & End-effector position and orientation, gripper state,
          and gripper opening. \\

          \textcolor{RBObsInk}{\texttt{freshness}}\newline
          \textcolor{RBObsInk}{\texttt{uncertainties}}
          & Observation validity and explicit caveats about
          missing or unreliable measurements. \\

          \bottomrule[0.8pt]
      \end{tabularx}
      \endgroup
  \end{table}

Action execution is synchronized with observation refresh. The planner loop permits one tool invocation per response, and the dispatcher serializes each physical action with publication of its resulting observation. An action attempt invalidates the previous frame. After the primitive returns, the adapter records its result and refreshes perception before the next action decision. In simulation, this refresh is attempted even when the primitive raises an error; unavailable observations are reported explicitly. A tool invocation can contain multiple low-level control steps, so this synchronization occurs at tool boundaries. The API planner retains text and tool history while keeping only the latest observation images in subsequent requests.

\section{Action Tools and Agent-Guided Execution}

The action library provides a common interface between the planner's subtask decisions and robot execution. It covers Cartesian motion, orientation adjustment, gripper control, and policy-based manipulation, as summarized in Table~\ref{tab:appendix-sim-tools}. The planner uses the
current observation and procedural skill to select an operation and express its intent through a position, a pose, a gripper command, or a language instruction. The environment adapter translates the request into executable robot actions. This organization lets skills describe how operations should be composed while delegating their low-level execution to the corresponding tools.

Geometric tools support operations with explicit spatial objectives,
including approaching an object, aligning the end effector, transporting a held object, and positioning it for release. Orientation and gripper tools provide complementary control over alignment and object handling. Policy-based tools supply learned manipulation behavior for grasping and other contact interactions. A task procedure can combine geometric positioning with policy-based manipulation, with the planner selecting the appropriate tool according to the current subtask and observed state.

For policy-based manipulation, the agent specifies the subtask instruction and determines whether additional inference-time guidance is needed. When guidance is activated, the agent configures a reward specification from the subtask objective, execution constraints, and observed feedback. This specification steers action generation toward the intended operation while keeping the policy parameters fixed. The policy produces short action sequences from current sensor observations, and the agent evaluates the resulting progress at tool boundaries to inform subsequent invocations. The policy thus supplies continuous manipulation capability, while the agent controls its invocation and the guidance appropriate to
the current operation.

\begin{table}[h]
\centering

\caption{Functional organization of the action library. Tool names describe
their roles in task execution. Geometric tools provide explicit motion and
gripper control, while policy-based tools support learned manipulation
with inference-time guidance configured by the agent when needed.}
\label{tab:appendix-sim-tools}

\vspace{5pt}

\begingroup

% Horizontal cell padding
\setlength{\tabcolsep}{6pt}

% Vertical cell padding
\renewcommand{\arraystretch}{1.35}

% Vertically center the automatically sized X column
\renewcommand{\tabularxcolumn}[1]{m{#1}}

% Keep outer padding so row colors align with horizontal rules.
% tabularx includes this padding in the total width.
\begin{tabularx}{\linewidth}{
    >{\raggedright\arraybackslash}m{0.22\linewidth}
    >{\raggedright\arraybackslash}m{0.24\linewidth}
    >{\raggedright\arraybackslash}X
}

% ================= Header =================
\arrayrulecolor{tableline}
\toprule

\rowcolor{tableheader}
\textbf{Function}
&
\textbf{Tool names}
&
\textbf{Role}
\\[3pt]

\midrule

% ================= Row 1 =================
Cartesian motion
&
\begin{tabular}[c]{@{}l@{}}
Cartesian Move\\
Pose Alignment
\end{tabular}
&
Approach, transfer, and positioning with optional orientation adjustment.
\\[3pt]

% ================= Row 2 =================
\rowcolor{tablerow}
Orientation adjustment
&
\begin{tabular}[c]{@{}l@{}}
Wrist Rotation\\
Wrist Tilt
\end{tabular}
&
Adjust end-effector orientation while maintaining its position.
\\[3pt]

% ================= Row 3 =================
Gripper control
&
\begin{tabular}[c]{@{}l@{}}
Open Gripper\\
Gripper Control
\end{tabular}
&
Open or actuate the gripper for object handling.
\\[3pt]

% ================= Row 4 =================
\rowcolor{tablerow}
\begin{tabular}[c]{@{}l@{}}
Policy-based\\
manipulation
\end{tabular}
&
\begin{tabular}[c]{@{}l@{}}
Policy Grasp\\
Policy Contact
\end{tabular}
&
Execute learned grasping and contact behaviors, with agent-configured
inference-time guidance when needed.
\\[3pt]

\bottomrule

\end{tabularx}

\endgroup
\end{table}

Tool execution is embedded in the observation--action--verification loop. An invocation returns execution feedback and triggers an observation refresh, allowing the planner to check whether the intended intermediate outcome has been achieved. The planner then continues the procedure, adjusts the next operation, or stops in response to the observed result. Each tool operates within its stopping conditions and execution budget. Tool completion and task success are assessed separately: a returned operation does not by itself establish that the task is complete. Simulation success is determined by the benchmark, while physical evaluation records an operator-provided final label.

Across environments, adapters preserve compatible operation semantics while accommodating differences in sensing, control, and tool availability. The physical adapter checks workspace and motion constraints before execution. Gripper operations associated with Cartesian motion follow verified arrival at the target, and policy-tool invocation requires successful positioning and a refreshed observation. Execution errors halt the physical episode. These conditions govern how tools can be composed on the robot while allowing the planner to reuse task procedures through the adapted interfaces.

\section{LIBERO-PRO Benchmark and Evaluation Protocol}
\label{app:liberopro-evaluation}

LIBERO-PRO augments LIBERO with controlled variations in task requirements and scene configurations. Our evaluation covers the SPATIAL, OBJECT, GOAL, and LIBERO-10 families, which provide complementary settings involving spatial relationships, object identities, goal conditions, and longer manipulation sequences. Together, these families allow us to examine procedural skills across different manipulation requirements under a common benchmark framework.

Each family is evaluated under two perturbation conditions, T and S. In the task or instruction-redirection condition T, the instruction specifies another valid target object or goal condition. The agent must therefore interpret the revised objective and bind its procedure to the appropriate objects. In the swap or position-swap condition S, object starting locations are exchanged or rearranged while the instruction is kept unchanged. This condition requires the agent to ground its actions in the current layout. The two conditions examine adaptation to changes in task meaning and spatial configuration, respectively.

For every task, seeds 11--20 are assigned to training for skill evolution, seeds 21--25 to validation, and seeds 1--10 to testing. Training episodes supply the execution feedback used to construct candidate skill revisions. The validation gate compares a candidate with the current skill under matched initial conditions and determines whether the revision is retained. The selected skill is subsequently fixed throughout testing. The three seed sets are disjoint, so test initial states are excluded from both skill-update evidence and validation decisions. Reported success rates are calculated from test episodes alone.

A test rollout starts from the initial state specified by its task and seed. RoboBridge receives the task instruction and initial observation, then uses the selected skill to organize tool execution and verify intermediate outcomes. Observations and execution feedback are refreshed as the rollout proceeds, while the skill itself remains unchanged. The rollout continues until the task succeeds, execution is stopped, or the allotted budget is exhausted. This distinction allows the planner to respond to the current scene during testing without performing further skill evolution.

The four families and two perturbation conditions form $4\times2=8$ evaluation groups. Each group contains $10$ tasks, with one test rollout for each of the $10$ test seeds per task. Its reported test count is therefore
\begin{equation}
    N_{\mathrm{group}} = N_{\mathrm{tasks}} \times N_{\mathrm{test\ seeds}} = 10 \times 10 = 100.
    \label{eq:liberopro-group-rollouts}
\end{equation}
Across all groups, each evaluated configuration covers $80$ task--perturbation combinations and
\begin{equation}
    N_{\mathrm{test}} = 4 \times 2 \times N_{\mathrm{group}} = 800
    \label{eq:liberopro-total-rollouts}
\end{equation}
test rollouts. Training episodes and validation rollouts are accounted for separately and are excluded from these test counts.

LIBERO-PRO's task-completion predicate supplies the success label for each test rollout. Let $y_{g,i,s}$ denote the binary outcome for task $i$ and test seed $s$ in evaluation group $g$. It equals $1$ if the official predicate is satisfied within the execution budget and $0$ otherwise. With $\mathcal{S}_{\mathrm{test}}=\{1,\ldots,10\}$, the success rate of group $g$ is
\begin{equation}
    \mathrm{SR}_{g} = \frac{1}{N_{\mathrm{group}}} \sum_{i=1}^{10} \sum_{s\in\mathcal{S}_{\mathrm{test}}} y_{g,i,s}.
    \label{eq:liberopro-success-rate}
\end{equation}
The same task-level criterion applies regardless of the sequence of tools used to execute the task. A primitive returning control to the planner, or the agent declaring completion, does not by itself establish success.

\section{VLA Model Instantiations}
\label{app:vla-models}

RoboBridge uses a frozen $\pi_{0.5}$ policy as its contact-rich action primitive in both simulation and physical deployment. The two deployments use different checkpoints because the simulated and real environments provide different robot embodiments, sensor layouts, state representations, and control interfaces. In simulation, we use the RLinf LIBERO checkpoint, denoted $\pi_{\mathrm{RLinf}}$, while the physical experiments use the OpenPI DROID checkpoint, denoted $\pi_{\mathrm{DROID}}$ \citep{intelligence2025pi_,yu2025rlinf}. During each policy invocation, the agent may construct an inference-time reward specification from the current subtask, execution constraints, and observed feedback. This reward guidance steers the generated action segment without updating the VLA parameters.

\subsection{$\pi_{\mathrm{RLinf}}$: RLinf LIBERO Checkpoint}
\label{app:pi-rlinf}

\paragraph{Architecture.}
$\pi_{\mathrm{RLinf}}$ is the LIBERO-specific $\pi_{0.5}$ checkpoint released with RLinf, corresponding to the full-shot \texttt{pi05-libero130} model. The policy combines visual observations, language instructions, and robot state before producing continuous robot actions. Within RoboBridge, the planner specifies the current operation through a task or subtask instruction, while the agent can supplement this context with an inference-time reward specification. The policy then converts the multimodal context and the guidance signal into a short manipulation segment.

\paragraph{Input, Guidance, and Action Interface.}
For LIBERO execution, the policy receives an external RGB view, a wrist-camera RGB view, proprioceptive state, and a language prompt. The environment adapter maps these fields to the input structure expected by the $\pi_{0.5}$ implementation and applies the normalization statistics associated with the checkpoint. Before a guided invocation, the agent interprets the active skill and execution feedback to define the desired local outcome, relevant constraints, and undesirable behaviors. This information is expressed as an inference-time reward specification and used to steer action generation during the current policy call. The resulting action chunk is then executed before the next observation is returned to the planner.

\paragraph{Action Modeling.}
The inspected LIBERO service produces a short chunk of continuous actions rather than a single control command. Each chunk contains seven-dimensional environment actions and is executed before the planner receives the refreshed observation and execution status. The reward guidance operates during this action-generation process: it evaluates whether candidate behavior is consistent with the local subtask and biases the generated segment toward the desired outcome. The guidance is therefore local to the current invocation and does not become a permanent modification of the policy.

\paragraph{Training and Deployment.}
The checkpoint is adopted without additional policy training or target-domain fine-tuning. Its parameters are loaded once and kept in evaluation mode. During guided inference, the policy parameters remain frozen; gradients, when
used, are computed only with respect to the candidate action sequence. The reward specification supplied by the agent is an inference-time control signal rather than a supervised training label or a parameter-update objective. Skill evolution consequently operates at the procedural level, modifying operation ordering, target binding, tool selection, and verification conditions, while reward-guided inference adjusts the execution of the selected operation.

\paragraph{Performance and Role in RoboBridge.}
When used directly as a policy-only baseline on LIBERO-PRO, $\pi_{\mathrm{RLinf}}$ obtains an overall success rate of $50.0\%$ under the reported evaluation protocol. This result provides a reference for separating the contribution of the frozen action policy from the contribution of procedural planning, skill evolution, and inference-time reward guidance. In RoboBridge, $\pi_{\mathrm{RLinf}}$ supplies fine-grained manipulation capability, while the agent determines the operation to execute, constructs guidance when necessary, and verifies whether the resulting state satisfies the intended subtask.

\subsection{$\pi_{\mathrm{DROID}}$: OpenPI DROID Checkpoint}
\label{app:pi-droid}

\paragraph{Architecture and Checkpoint.}
$\pi_{\mathrm{DROID}}$ is the OpenPI $\pi_{0.5}$ checkpoint used for physical-robot execution. It preserves the same general vision-language-action formulation as the LIBERO policy, but is supplied through a DROID-oriented deployment stack. The physical agent does not invoke the checkpoint with a fixed open-loop procedure alone. Instead, it provides the current manipulation instruction together with an inference-time reward specification when additional behavioral steering is required. The checkpoint remains unchanged while the guidance signal adapts action generation to the current physical context.

\paragraph{Observation, Guidance, and Action Processing.}
The physical policy service consumes an external view, a wrist view, robot state, and a language instruction describing the current manipulation operation. The agent first obtains a fresh observation and positions the end effector for the intended operation. It then determines whether the policy should be guided and, if so, specifies the desired interaction outcome and physical constraints through the reward signal. The DROID policy generates a short action chunk under this instruction and guidance, which is executed through the physical control interface. After execution, refreshed images and robot state are returned for verification before another policy call is permitted.

\paragraph{Frozen Use and Agent Coordination.}
The DROID checkpoint remains frozen during real-world skill evolution. The agent controls the conditions under which the policy is invoked: it selects the target object, determines the local operation, checks that the preceding positioning action succeeded, and constructs the reward specification when the action requires additional steering. The reward signal is used only during inference and does not create a new training update for the policy. Subsequent skill revisions operate on the surrounding procedure, including pre-positioning, action ordering, recovery behavior, and outcome verification.

\paragraph{Role in Physical Transfer.}
In the real-robot stage, $\pi_{\mathrm{DROID}}$ provides the low-level manipulation capability required to execute transferred procedures. RoboBridge transfers the task-level organization of the skill from simulation, while the physical adapter and the agent-generated reward guidance determine how each local operation is realized under the target robot's observations and dynamics. Execution feedback is retained for subsequent skill revision, allowing the transferred procedure to be refined without retraining the entire VLA model. The contribution of $\pi_{\mathrm{DROID}}$ is therefore evaluated as part of the complete real-robot transfer and adaptation process rather than as an isolated policy-only score.

\section{Prompt, Skill, and Structured Artifact Organization}
\label{app:prompt-skill-artifacts}

RoboBridge separates agent behavior into three persistent layers: environment-specific system prompts, executable procedural skills, and structured execution artifacts. The system prompts define how the agent interprets observations, selects tools, and verifies outcomes. Skills describe reusable task procedures and their recovery conditions. Structured artifacts record the observations, tool calls, feedback, and evaluation results generated during an episode. The following examples present the organization and semantics of these components while omitting implementation-specific prompt text.

\subsection{Simulation Planner Prompt}

The simulation system prompt defines the common planner contract used during LIBERO-PRO execution. It instructs the agent to reason from the newest observation, issue at most one physical operation per response, and wait for the resulting observation before selecting the next operation. It also describes the available geometric and policy-based tools, the distinction between primitive completion and task success, and the requirement to use the benchmark completion predicate for final evaluation. The prompt does not encode a complete task procedure; task-specific behavior is supplied through the active skill.

\begin{tcolorbox}[
  enhanced,
  colback=white,
  colframe=black,
  colbacktitle=black,
  fonttitle=\bfseries\itshape\color{white},
  title={Prompt 1: Simulation Planner System Prompt},
  arc=3pt,
  boxrule=1pt,
  breakable
]

\textbf{Purpose:} Organize task execution in the simulator.

\vspace{0.4em}
\noindent\colorbox{blue!10}{
\parbox{\dimexpr\linewidth-2\fboxsep}{
\textbf{Core semantics:} Use the newest observation; select one operation;
wait for execution feedback; verify the resulting state.
}}

\vspace{0.2em}
\noindent\colorbox{gray!10}{
\parbox{\dimexpr\linewidth-2\fboxsep}{
\textbf{Tool semantics:} Combine geometric motion tools with policy-based
manipulation tools. Tool completion is not equivalent to task success.
}}

\vspace{0.2em}
\noindent\colorbox{green!10}{
\parbox{\dimexpr\linewidth-2\fboxsep}{
\textbf{Evaluation rule:} Use the benchmark task-completion predicate
to determine the final outcome.
}}

\end{tcolorbox}

\subsection{Physical-Robot System Prompt}

The physical-robot system prompt extends the simulation contract with requirements imposed by hardware execution. It requires the agent to inspect a fresh external and wrist observation before acting, respect workspace and motion constraints, and stop when execution or sensing becomes unreliable. A policy-based action call is allowed only after the robot has been positioned for the intended operation and a new observation has been obtained. The prompt also makes clear that closing the gripper, completing an action chunk, or returning from a tool does not prove that the intended object interaction succeeded.

\begin{tcolorbox}[
  enhanced,
  colback=white,
  colframe=black,
  colbacktitle=black,
  fonttitle=\bfseries\itshape\color{white},
  title={Prompt 2: Physical-Robot Execution Prompt},
  arc=3pt,
  boxrule=1pt,
  breakable
]

\textbf{Purpose:} Add hardware-aware constraints to the planner loop.

\vspace{0.4em}
\noindent\colorbox{blue!10}{
\parbox{\dimexpr\linewidth-2\fboxsep}{
\textbf{Observation rule:} Inspect the newest external and wrist views
before making a physical decision.
}}

\vspace{0.2em}
\noindent\colorbox{orange!12}{
\parbox{\dimexpr\linewidth-2\fboxsep}{
\textbf{Policy-call gate:} Position the end effector successfully and
refresh the observation before invoking the policy action.
}}

\vspace{0.2em}
\noindent\colorbox{red!10}{
\parbox{\dimexpr\linewidth-2\fboxsep}{
\textbf{Failure rule:} Stop the episode when execution, localization, or
observation reliability is lost.
}}

\end{tcolorbox}

\subsection{Skill-Evolution System Prompt}

The skill-evolution system prompt is used after one or more training rollouts. It asks the agent to compare the current skill with the observed execution evidence, identify the local cause of failure or improvement, and propose a bounded revision. Candidate revisions may modify an operation, transition, recovery condition, or verification rule while preserving unaffected parts of the skill. The prompt also requires the candidate to be evaluated against the incumbent skill before persistence. A candidate that does not improve execution or introduces unsupported behavior is rejected.

\begin{tcolorbox}[
  enhanced,
  colback=white,
  colframe=black,
  colbacktitle=black,
  fonttitle=\bfseries\itshape\color{white},
  title={Prompt 3: Skill-Evolution System Prompt},
  arc=3pt,
  boxrule=1pt,
  breakable
]

\textbf{Purpose:} Convert rollout evidence into a controlled skill revision.

\vspace{0.4em}
\noindent\colorbox{blue!10}{
\parbox{\dimexpr\linewidth-2\fboxsep}{
\textbf{Evidence:} Read observations, tool outcomes, intermediate feedback,
and the final task result.
}}

\vspace{0.2em}
\noindent\colorbox{yellow!15}{
\parbox{\dimexpr\linewidth-2\fboxsep}{
\textbf{Revision scope:} Add, replace, or remove a local operation,
transition, recovery condition, or verification rule.
}}

\vspace{0.2em}
\noindent\colorbox{green!10}{
\parbox{\dimexpr\linewidth-2\fboxsep}{
\textbf{Persistence rule:} Retain a candidate only after comparison with
the current skill under matched evaluation conditions.
}}

\end{tcolorbox}

\subsection{Stored Skill Structure}

A stored skill is represented as a compact procedural document rather than as a raw transcript. It contains the task intent, applicable conditions, expected observations, ordered operations, verification criteria, and recovery behavior. The skill is written around task-relevant states and affordances, so that object names or scene-specific coordinates do not become the only source of generalization. The following schematic shows the organization of a saved skill without reproducing a complete task-specific procedure.

\begin{tcolorbox}[
  enhanced,
  colback=white,
  colframe=black,
  colbacktitle=black,
  fonttitle=\bfseries\itshape\color{white},
  title={Artifact 1: Stored Skill Schema},
  arc=3pt,
  boxrule=1pt,
  breakable
]

\textbf{Skill ID:} \texttt{object\_transfer}

\vspace{0.3em}
\textbf{Intent:} Transfer a selected object to its designated destination.

\vspace{0.3em}
\textbf{Preconditions:}
\begin{itemize}
    \item The target and destination are observable.
    \item The robot is in a reachable starting configuration.
\end{itemize}

\textbf{Procedure:}
\begin{enumerate}
    \item Bind the instruction to the current target and destination.
    \item Select a suitable positioning or manipulation operation.
    \item Execute the local action and refresh the observation.
    \item Verify the intermediate state before continuing.
\end{enumerate}

\textbf{Verification:} Check target identity, relative placement, and task progress.

\vspace{0.3em}
\textbf{Recovery:} Re-observe, revise the local operation, or terminate when
the required evidence is unavailable.

\end{tcolorbox}

\subsection{Skill Versioning and Validation}

Each accepted skill is associated with a version identifier and the evidence used to produce it. During evolution, the incumbent skill remains available as a fallback, while a candidate skill is evaluated separately. The candidate is promoted only when it satisfies the validity checks and improves the selected execution criterion. Rejected candidates are retained as audit records rather than merged into the active skill. This versioned organization prevents a single failed rollout from permanently altering the procedural memory.

\begin{tcolorbox}[
  enhanced,
  colback=white,
  colframe=black,
  colbacktitle=black,
  fonttitle=\bfseries\itshape\color{white},
  title={Artifact 2: Skill-Update Record},
  arc=3pt,
  boxrule=1pt,
  breakable
]

\textbf{Current skill:} \texttt{skill\_v3}

\textbf{Candidate skill:} \texttt{skill\_v4-candidate}

\textbf{Evidence summary:} Repeated failure at the local transfer step;
preceding positioning operation remains valid.

\textbf{Proposed change:} Replace the local transfer transition and add
an intermediate verification condition.

\textbf{Validation result:} Candidate compared with the current skill
under the designated validation episodes.

\textbf{Decision:} \texttt{accepted} or \texttt{rejected}

\end{tcolorbox}

\subsection{Structured Observation and Episode Files}

The execution record is divided into files with different responsibilities. The skill file stores procedural knowledge, the observation snapshot stores the agent-visible scene state, the audit file stores perception and inventory information, and the rollout record stores tool calls, feedback, and the final evaluation result. This separation keeps internal perception details out of the planner request while preserving them for later diagnosis. A schematic observation record is shown below.

\begin{tcolorbox}[
  enhanced,
  colback=white,
  colframe=black,
  colbacktitle=black,
  fonttitle=\bfseries\itshape\color{white},
  title={Artifact 3: Observation Snapshot},
  arc=3pt,
  boxrule=1pt,
  breakable
]

\texttt{\{}

\begin{quote}
\texttt{"observation\_id": "obs\_k",}\\
\texttt{"episode\_step": "k",}\\
\texttt{"coordinate\_frame": "robot\_base",}\\
\texttt{"images": ["external\_rgb", "wrist\_rgb"],}\\
\texttt{"entities": [}\\
\hspace*{1em}\texttt{\{"id": "entity\_1", "measurement": "available"\}}\\
\texttt{],}\\
\texttt{"robot": \{"state": "current"\},}\\
\texttt{"freshness": "latest"},\\
\texttt{"uncertainties": ["target\_depth"]}\\
\end{quote}

\texttt{\}}

\vspace{0.3em}
\textbf{Note:} The example shows the schema only; implementation-specific
coordinates, detector outputs, and internal metadata are omitted.

\end{tcolorbox}

\subsection{Episode Artifact Organization}

A completed episode contains the files needed to reconstruct the interaction without treating the internal prompt transcript as the skill itself. A typical record includes the active skill version, step-indexed observations, image references, tool calls, execution feedback, perception audit information, and the final evaluation label. The stored files support both failure analysis and skill evolution: the agent can identify whether a failure originated from object binding, positioning, action generation, tool ordering, or outcome verification.

\begin{tcolorbox}[
  enhanced,
  colback=white,
  colframe=black,
  colbacktitle=black,
  fonttitle=\bfseries\itshape\color{white},
  title={Artifact 4: Schematic Episode Record},
  arc=3pt,
  boxrule=1pt,
  breakable
]

\texttt{episode/}

\hspace*{1em}\texttt{skill.md} \hfill active procedural skill

\hspace*{1em}\texttt{snapshots/} \hfill step-indexed observations

\hspace*{1em}\texttt{images/} \hfill external and wrist views

\hspace*{1em}\texttt{transcript.json} \hfill planner requests and tool calls

\hspace*{1em}\texttt{audit.json} \hfill perception and inventory details

\hspace*{1em}\texttt{rollout.json} \hfill execution feedback and outcome

\hspace*{1em}\texttt{operator\_label.json} \hfill physical evaluation label

\end{tcolorbox}

\subsection{Simulation and Real-Robot Prompt Alignment}

The simulation and physical prompts share the same high-level behavioral contract but differ in their environment-specific constraints. The simulation prompt emphasizes benchmark interaction and predicate-based completion, whereas the physical prompt additionally enforces fresh sensing, pre-positioning before policy invocation, hardware-aware stopping, and operator-based outcome labeling. The skill format and update protocol remain shared across the two domains. This organization allows a skill evolved in simulation to be transferred as a procedural object while adapting its environment-specific execution conditions through the physical prompt and feedback records.

\end{document}